\documentclass[conference]{IEEEtran}
\IEEEoverridecommandlockouts
\usepackage{cite}
\usepackage{amsmath,amssymb,amsfonts}
\usepackage{algorithmic}
\usepackage{graphicx}
\usepackage{textcomp}
\usepackage{xcolor}
\def\BibTeX{{\rm B\kern-.05em{\sc i\kern-.025em b}\kern-.08em
    T\kern-.1667em\lower.7ex\hbox{E}\kern-.125emX}}
\begin{document}

\title{Federated Learning of Low-Rank One-Shot Image Detection Models in Edge Devices with Scalable Accuracy and Compute Complexity\\
}

\author{\IEEEauthorblockN{Abdul Hannaan$^1$, Zubair Shah$^1$, Aiman Erbad$^2$, Amr Mohamed$^2$, Ali Safa$^1$}
\IEEEauthorblockA{$^1$\textit{College of Science and Engineering, Hamad Bin Khalifa University, Doha Qatar} \\
$^2$\textit{College of Engineering, Qatar University, Doha Qatar} \\
abdulhannaan2002@gmail.com , asafa@hbku.edu.qa}
}

\linespread{0.87}

\maketitle

\begin{abstract}
This paper introduces a novel federated learning framework termed LoRa-FL designed for training low-rank one-shot image detection models deployed on edge devices. By incorporating low-rank adaptation techniques into one-shot detection architectures, our method significantly reduces both computational and communication overhead while maintaining scalable accuracy. The proposed framework leverages federated learning to collaboratively train lightweight image recognition models, enabling rapid adaptation and efficient deployment across heterogeneous, resource-constrained devices. Experimental evaluations on the MNIST and CIFAR10 benchmark datasets, both in an independent-and-identically-distributed (IID) and non-IID setting, demonstrate that our approach achieves competitive detection performance while significantly reducing communication bandwidth and compute complexity.  This makes it a promising solution for adaptively reducing the communication and compute power overheads, while not sacrificing model accuracy. 
\end{abstract}


\section*{Supplementary Material}

Code is released at 
\texttt{https://tinyurl.com/mp3fsyk9}

\section{Introduction}

The rapid proliferation of edge devices has created a pressing need for efficient and accurate image detection models that can operate in real-time under stringent resource constraints \cite{intr2, intr1}. Traditional Deep Learning models, while highly accurate, often require extensive computational resources \cite{edgeeff1} and large volumes of labeled data—resources that are rarely available in decentralized, edge-based scenarios \cite{aiman1}. At the same time, \textit{Federated learning} (FL) \cite{b1} has emerged as a viable paradigm to overcome these challenges by enabling multiple edge devices to collaboratively train models without sharing raw data, thereby preserving privacy and reducing central storage demands \cite{aiman1, aiman2, medicalfl, privacy}.

\begin{figure}[htbp]
    \centering
    \includegraphics[width=0.47\textwidth]{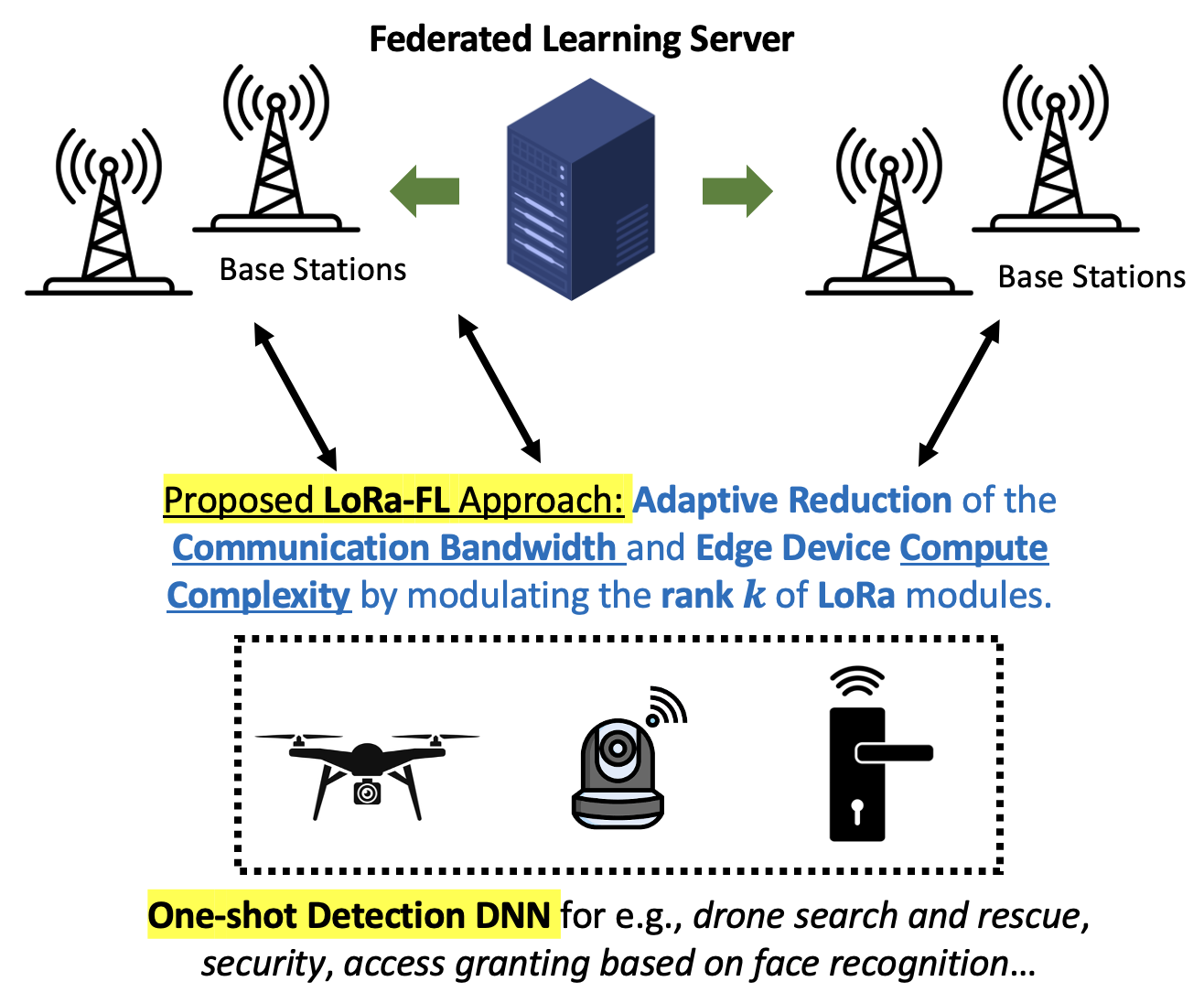} 
    \caption{\textit{\textbf{Scenario considered in this work.} A number of edge devices with limited compute capacity are considered within a Federated Learning (FL) setting, where the goal is to learn a one-shot image detection model in a federated manner. Given a target image, the one-shot model can predict if an incoming query image is of the same class as the target or not. Within this context, we propose an approach termed LoRa-FL for adaptively reducing the edge-server communication bandwidth requirements as well as the compute complexity needed for the edge devices during the FL process. Our approach makes use of Low-Rank (LoRa) modules within the one-shot network architecture which effectively reduces the number of trainable weights by modulating the rank $k$ of the LoRa modules.}}
    \label{graphasbt}
\end{figure}

However, deploying state-of-the-art image detection models in a federated setting presents its own set of challenges. In particular, the high communication and computation costs associated with full-model updates can significantly hinder the scalability and responsiveness of FL systems on edge devices \cite{energ2}. Moreover, the inherent heterogeneity in edge environments, ranging from varying bandwidth and compute power, to non-independent-and-identically-distributed (non-IID) data sources, further complicates the training process \cite{noniidamr}.

To help alleviate the compute and memory cost of training Deep Neural Networks (DNNs), \textit{Low‑Rank Adaptation} (LoRA) has been introduced as an efficient fine‑tuning technique that leverages low‑dimensional parameter updates during the training of deep learning models \cite{b2}. By updating weight matrices or low rank $k$, LoRA not only reduces the overall memory and computation footprint, but can also minimize the volume of data exchanged during model updates. Hence, integrating LoRA into the federated learning framework holds great promise: it can substantially alleviate the high resource costs associated with full‑model updates while enabling rapid model adaptation. 

In this paper, we propose a novel federated learning framework that integrates low-rank adaptation into \textit{one-shot image detection} models. We apply our proposed low-rank FL approach to the problem of \textit{one-shot image recognition}, where the goal is to recognize an incoming query image with regard to a \textit{single} target training image—a scenario that closely aligns with a plethora of possible edge AI applications, from one-shot drone visual place recognition \cite{drone1,drone2} and user face recognition to one-shot object detection and tracking \cite{b4} (see Fig. \ref{graphasbt}). Our approach is built on the observation that many deep learning models exhibit significant redundancy, allowing their weight parameters to be approximated by low-rank matrices. By updating only these low-dimensional representations during federated training, our method adaptively reduces \textit{i)} the edge-server communication bandwidth by reducing the number of trained parameters; and \textit{ii)} the local compute complexity required in the edge device while preserving the core detection capabilities of the model.

The contributions of this paper are as follows:
\begin{enumerate}
    \item We introduce a method which fuses Federated Learning with the use of Low Rank (LoRa) weight adapters for the training of one-shot image detection models embedded within a number of edge devices, which communicate with a central federation server.
    \item By conducting numerous experiments, we show that our proposed LoRa-FL approach enables a control of the trade-off between model accuracy, communication bandwidth and edge device compute complexity, making it a highly attractive method for the design of efficient federated learning systems.
    \item We release our code as open-source for helping future research in the field.
\end{enumerate}

\begin{figure*}[htbp]
    \centering
    \includegraphics[scale=0.4]{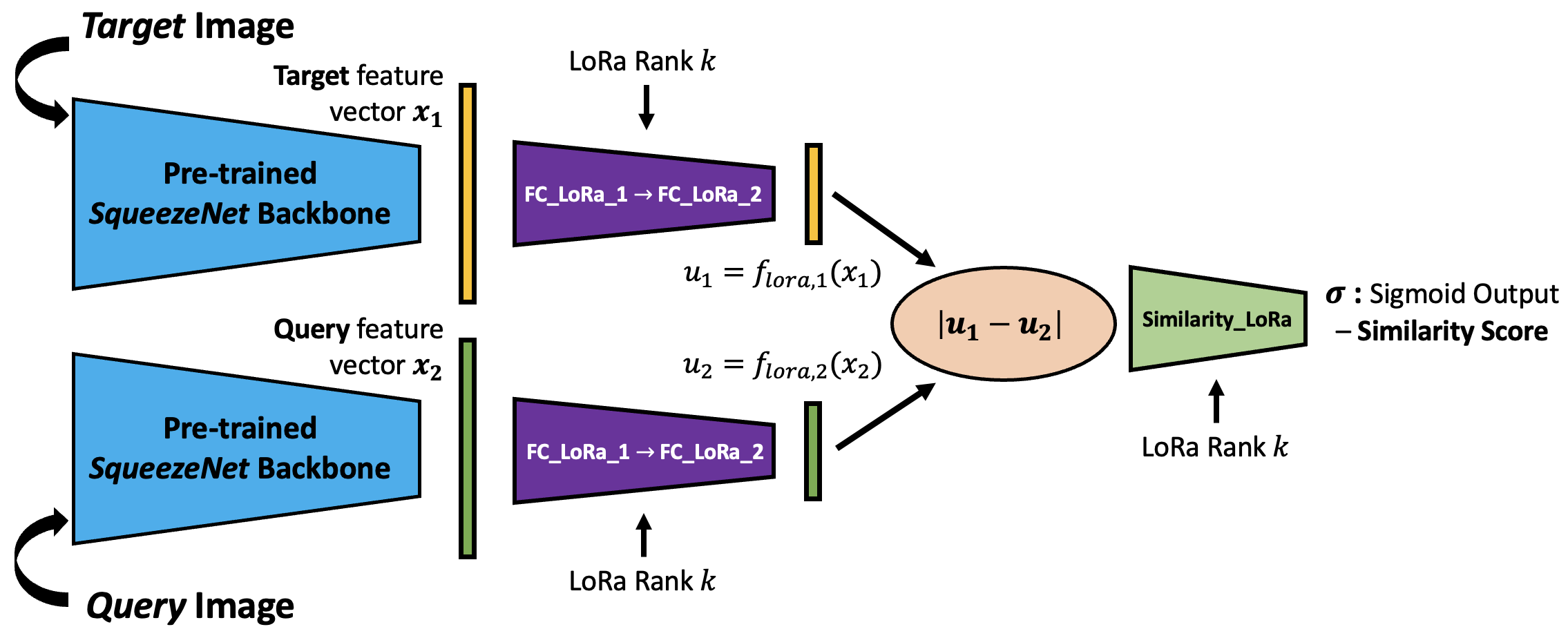} 
    \caption{\textit{\textbf{One-shot Siamese model architecture considered in this work.} The target and query images are fed to their respective pre-trained SqueezeNet backbone, acting as a Siamese feature extraction module. The target and query feature vectors $x_1$ and $x_2$ are then processed by two cascaded LoRa layers of rank $k$ (where $k$ varies during our experiments). The produced target and query embeddings $u_1$ and $u_2$ are then used to compute the absolute difference vector $|u_1 - u_2|$ which is in turn processed by a final LoRa module. Finally, the output of the \texttt{similarity\_LoRa} module is processed by a Sigmoid layer with one output indicating similarity or dissimilarity between the target and query images.}}
    \label{fig1}
\end{figure*}

The remainder of this paper is organized as follows. Related work are reviewed in Section \ref{relatedwork}. Our proposed approach is presented in Section \ref{proposedapproach}. Experimental results are provided in Section \ref{experim}. Finally, conclusions are provided in Section \ref{conc}.

\section{Related Work}
\label{relatedwork}

To the best of our knowledge, most prior work related to the use of low-complexity model training techniques such as LoRa have been carried in the context of Large Language Models (LLMs), without addressing the case of compact one-shot image detection models (which is the focus of this paper) \cite{b3, b5, b7, b6}. The study of various Parameter-Efficient Fine-Tuning (PEFT) techniques has recently gained attention in FL for mitigating the communication costs of LLMs. Z. Zhang \textit{et al.} \cite{b7} proposed a method termed FedPETuning which effectively applies LoRa adaptation on top of \textit{all} pre-trained weights of LLM models. In their case, LoRa adaptation is not used as the sole weight parameters within the network but rather as additional trainable parameters $\Delta W$ which add their contribution on top of pre-trained weights $W$ as $W_{tot}=W+\Delta W$. Then, the authors in \cite{b7} propose to train the obtained model within a FL context. They show that FedPETuning can maintain accuracy and even uphold privacy compared to full-model training, while significantly reducing the required communication bandwidth \cite{b7}. In another work by G. Sun \textit{et al.} \cite{b6}, the authors introduce a framework termed FedPEFT which combines various different techniques for low-complexity LLM fine-tuning under a single framework. These techniques include bias-only tuning (where the biases of each weight matrices are trained only), prefix tuning (which tunes a subset of the LLM layers only) and LoRa adaptation (where all weights are augmented with LoRa adapters as $W_{tot}=W+\Delta W$). By adaptively switching between these different low-complexity tuning techniques within an FL context, the authors in \cite{b6} show that their framework enables a reduction in total communication overheads by several orders of magnitude, while maintaining competitive (or even better) LLM accuracy across various FL settings \cite{b6}. Overall, these studies within the LLM context established that transmitting only a small, trainable subset of model parameters (as in LoRA or related methods) is a viable strategy to overcome FL’s communication bottleneck without heavy computation on clients.

However, these works were predominantly conducted in the context of LLMs with the aim of fine tuning very deep LLM backbone networks, which incurs significant computational costs at the edge device side. In addition, the application of LoRa in their LLM context is done by \textit{augmenting} the existing LLM weights $W$ with LoRa adapters $\Delta W$, as $W_{tot}=W+\Delta W$. During edge training, this makes the required forward pass go through many multiply-accumulate operations due to the application of both $W$ and $\Delta W$, jeopardizing memory and compute complexity at the edge \cite{compression}.

In contrast, our approach provides one of the first studies on the application of parameter-efficient tuning techniques for the case of \textit{one-shot image detection models}, as opposed to the LLMs that have been predominantly explored in prior work. Our LoRa-FL framework deviates from prior work by proposing to tune the output layers of a Siamese network architecture only. In addition, we do not use LoRa adapters as an augmentation method on already pre-trained output weight matrices (i.e., $W_{tot}=W+\Delta W$). Rather, we consider that the LoRa modules are the \textit{sole weights} of the model output layer (i.e., $W_{tot}=\Delta W$). Hence, by freezing the pre-trained backbone weights and updating only the low‑rank adapter parameters, our method significantly reduces both the number of parameters that need to be communicated and the computational load on each client. 

\section{Proposed Approach}
\label{proposedapproach}
Our proposed approach leverages FL for training one-shot image detection models using LoRA adapters as the sole weights to be trained in the model. 
Below, we provide an overview of the DNN architecture and the FL-LoRa training approach proposed in this work.

\subsection{One-shot Model Architecture}
The DNN model considered in this work flow is the  popular Siamese Network approach for one-shot learning \cite{siamese}, where the target and query images are processed with the same backbone network before being processed by a number of fully-connected layers that performs similarity scoring between the target and the query \cite{similaritysco}. Our architecture is shown in Fig. \ref{fig1} and begins with a pair of pre-trained SqueezeNet backbones \cite{squeezenet} (pre-trained using the ImageNet dataset \cite{imagenet}), which act as a feature extractors for both the target and query images. Choosing a SqueezeNet backbone is well-suited within our FL context since this architecture has been proposed for scenarios where compute power is restricted, as in the case of edge devices \cite{droneexp}. This backbone is not trained during the FL process to reduce the communication and computational load on the edge devices. At the output of the two Siamese backbones, we insert LoRA adapter layers (i.e., low-rank weight layers) that adapt the extracted features for performing similarity scoring as follows:
\vspace{-4pt}
\begin{equation}
\text{sim}(x_1,x_2) = \sigma\Bigl(\text{similarity\_lora}\Bigl(|u_1-u_2|\Bigr)\Bigr),
\label{simeq}
\end{equation}
where \( \sigma(\cdot) \) is the sigmoid activation function, ensuring outputs in the range \([0,1]\) ($0$ indicating non-similarity and $1$ indicating similarity between the target and the query image). Furthermore, $u_1$ and $u_2$ in (\ref{simeq}) are embeddings obtained by applying LoRa adapter layers (\texttt{fc\_lora\_1} and \texttt{fc\_lora\_2}) to the outputs $x_1$ and $x_2$ of the Siamese SqueezeNet backbones, producing embeddings (of dimensions 128) for both the target and query images: \( u_1 = f_{lora,1}(x_1) \) and \( u_2 = f_{lora,2}(x_2) \). Finally, $\text{similarity\_lora}$ in (\ref{simeq}) denotes the application of a LoRa weight module on the result of the absolute difference $|u_1-u_2|$, which produces a final scalar similarity score given the absolute difference between the target and query embeddings (\ref{simeq}). This design ensures that only a small fraction of the total parameters (i.e., LoRA adapters) are trained, minimizing both computation and communication overhead on edge devices. 


As loss function $\mathcal{L}$, we use the binary cross-entropy loss (BCE) \cite{bceloss} between the similarity output of the network (\ref{simeq}) and the labels from the one-shot datasets used in this work (see Section \ref{datasetconstr}):
\begin{equation}
    \mathcal{L} = \frac{1}{n}\sum_{j=0}^{n-1} y_j \log (\text{sim}(x_1,x_2)_j) + (1 - y_j) \log (1 - \text{sim}(x_1,x_2)_j)
\end{equation}
where $n$ denotes the size of the mini-batch (set to 32 during training). In addition, we make use of the Adam optimizer \cite{adam} with its default parameters during training, with a learning rate of $\eta = 10^{-3}$.

\subsection{One-shot Dataset Aggregation}
\label{datasetconstr}
The one-shot image recognition problem is treated as a \textit{similarity learning} task \cite{siamese}: each data sample within the training set is constructed as a pair of images from either the MNIST \cite{mnsitdataset} or the CIFAR10 \cite{cifar10} datasets, with a binary label indicating whether they belong to the same class or not (\textit{positive} or \textit{negative} pairs). The similarity learning dataset is constructed in a balanced way, with 50\% positive and 50\% negative pairs. Finally, the obtained one-shot dataset is split into a training set and a test set in order to enable the evaluation of the network accuracy during our experiments in Section \ref{experim}. 

During our experiments, we investigate both the IID and the non-IID cases for both the MNIST and the CIFAR10 datasets. Under the IID case, each client receives an approximately uniform image pair distribution while in the non-IID case, we group image classes into disjoint sets so that each client has access to certain specific classes only. This latter setting is significantly more challenging than the IID case due to \textit{catastrophic interference} phenomenon \cite{catastrophic}, but represents a more realistic case in situations where the edge devices do not have all access to the same data distributions due to the heterogeneity in their respective environments \cite{noniidamr}.

\subsection{Federated Learning - LoRa Pipeline}

Our FL pipeline follows the popular approach proposed in \cite{b1} alternating a \textit{Local Training} step at the client side and an \textit{Aggregation} step at the central server side, but deviates from traditional FL through the introduction of LoRa weight modules with varying precision rank $k$. During the \textit{Local Training} step, in each global round, a subset of clients train their one-shot DNN network on their local data. Since the backbone weights are not trained, only the LoRA layers are trained and updated. Then, during the \textit{Aggregation} step, the central server receives the client updates and merges these updates using the \texttt{FederatedAveraging} algorithm in \cite{b1}, by aggregating the LoRA adapter parameters only, while not modifying the weights in the Siamese SqueezeNet backbones. 

For each client $i$, LoRa modules essentially model a standard fully-connected weight matrix $W_i$ of size $(N,M)$ as a matrix-product between two low-rank matrices $A$ and $B$ of size $(N,k)$ and $(k,M)$ respectively, with rank $k$ as a precision-defining parameter:
\begin{equation}
W_i = A_i . B_i.
\end{equation}

Hence, during the FL procedure, a potentially much smaller amount of weights corresponding to the entries of $A_i$ and $B_i$ are effectively updated, reducing both the edge-server communication bandwidth and the compute complexity at the edge device side.

\section{Experimental results}
\label{experim}
In our experiments, we evaluated the performance of our LoRA-based federated learning approach using Siamese networks on MNIST \cite{mnsitdataset} and CIFAR-10 \cite{cifar10} datasets under both IID and Non-IID data distributions. The MNIST dataset \cite{mnsitdataset} consists of images of gray-scale hand-written digits (10 classes in total) while the more challenging CIFAR-10 dataset \cite{cifar10} consists of RGB images of objects such as planes, birds and so on (10 different object classes in total).   
The FL configuration consists of $C=5$ clients, with $K=3$ clients randomly selected to participate within each global round. Training was conducted over $10$ global rounds with each client performing local training for $5$ epochs. This setup was designed so as to simulate a scenario where a limited number of edge devices need to quickly learn within $10$ rounds with limited local training burden ($5$ epochs only). All our experiments are conducted following a $5$-fold accuracy assessment approach, where the network is trained and tested $5$ times with different random initialization and train-test split for each fold. Then, the final accuracy is reported as the average over the 5 folds.

During all our experiments below, we systematically vary the rank $k$ of the LoRa weight modules from $k=1$ to $k=32$ in order to study the impact of weight complexity on both FL convergence and model accuracy.

\subsection{MNIST - IID}
\label{mnistiid}
\begin{figure}[htbp]
    \centering
    \includegraphics[width=0.5\textwidth]{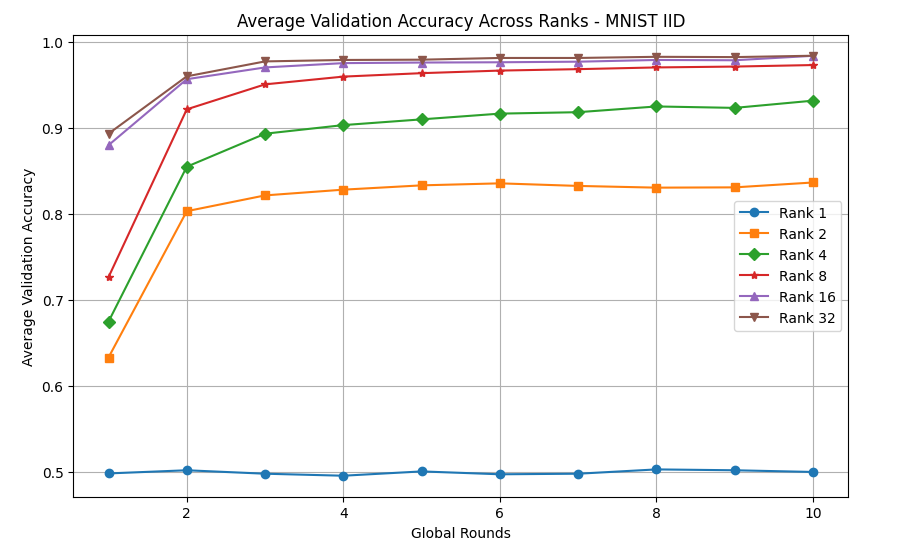} 
    \caption{\textit{\textbf{MNIST IID:} Test accuracy in function of the global FL round.}}
    \label{fig2}
\end{figure}

For the MNIST dataset under the IID setting, Fig. \ref{fig2} shows that the higher the rank $k$, the faster FL convergence is attained and the higher the test accuracy becomes, reaching over $98\%$ for $k\geq 8$. This is expected since the larger $k$ becomes, the more the LoRa readout layers within the one-shot learning network will be precise, leading to better test performance. On the other hand, the higher $k$ becomes, the more the compute complexity and communication bandwidth increases since more weights need to be trained (see Section \ref{computecomp}). It is interesting to see in Fig. \ref{fig2} that for $k\geq 8$, the final test accuracies become very similar, suggesting that $k=8$ is sufficient for attaining both high accuracy while minimizing communication and edge compute resources.

\subsection{CIFAR10 - IID}
\label{cifariid}
\begin{figure}[htbp]
    \centering
    \includegraphics[width=0.5\textwidth]{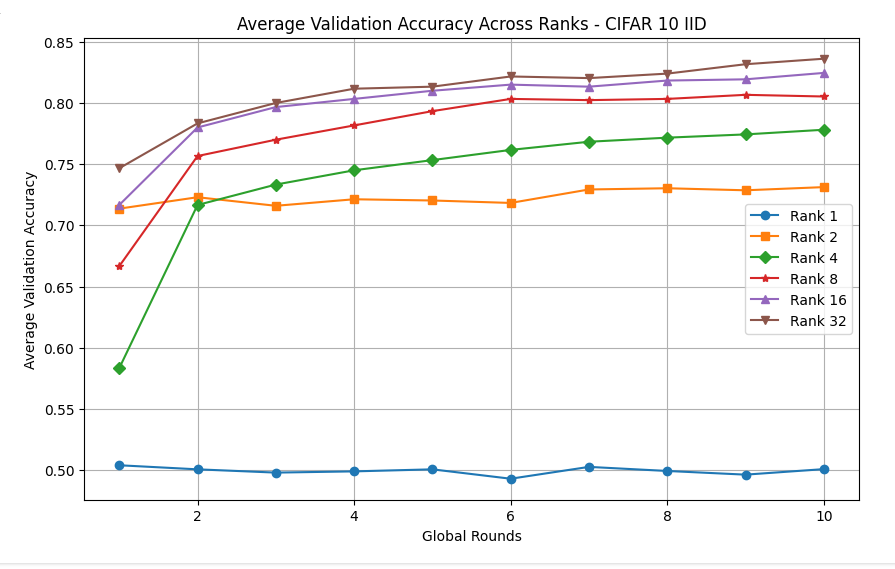} 
    \caption{\textit{\textbf{CIFAR10 IID:} Test accuracy in function of the global FL round.}}
    \label{fig4}
\end{figure}

For the more challenging CIFAR10 dataset under the IID setting, the differences between low and higher LoRA ranks $k$ were more significantly pronounced compared to the MNIST-IID case. Fig. \ref{fig4} shows that models with low ranks $k$ struggle to attain usable accuracies (e.g., above $75\%$), whereas for $k\geq8$ a steep increase in accuracy is observed, approaching 80–85\% within 4-6 global rounds.

\subsection{MNIST - Non-IID}

\begin{figure}[htbp]
    \centering
    \includegraphics[width=0.5\textwidth]{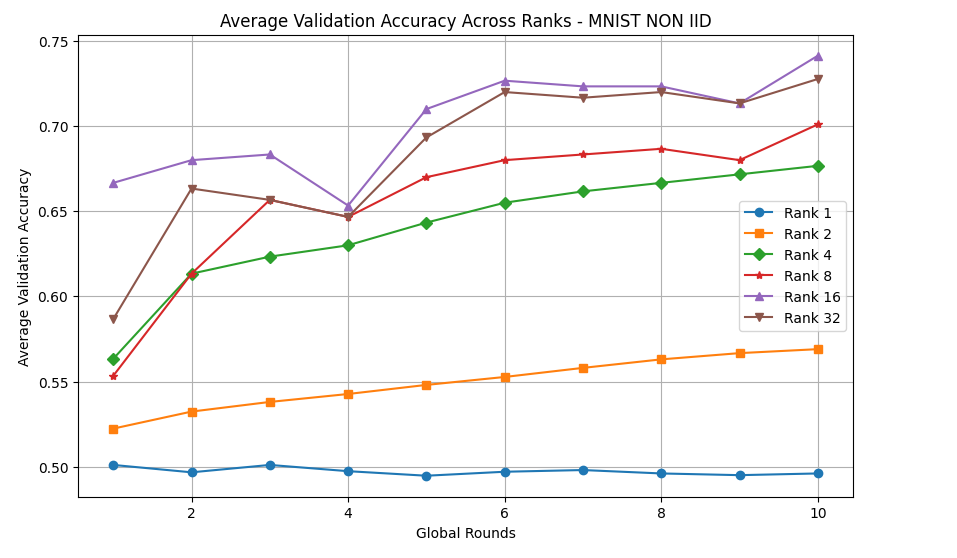} 
    \caption{\textit{\textbf{MNIST NON-IID:} Test accuracy in function of the global FL round.}}
    \label{fig3}
\end{figure}

We now study the effect of the LoRa rank $k$ under the non-IID data distribution setting, where each client is assigned with a subset of the image classes only (i.e., clients do not have access to all image classes but rather, client $1$ is assigned with class 1 and 2, client $2$ with class 3 and 4 etc.). This setting represents a much more challenging scenario compared to the IID case due to the \textit{catastrophic interference} phenomenon \cite{catastrophic} that is well-known to degrade model performance in non-IID datasets \cite{noniidamr}. Fig. \ref{fig3} shows the results obtained in the non-IID MNIST case. Compared to the IID case in Section \ref{mnistiid}, we observe an expected performance drop in terms of overall accuracy; however, even in these challenging non-IID conditions, we observe a convergence of the network test accuracy as the rank $k$ is increased further. Again, this indicate that using maximum rank is potentially not necessary beyond a certain point (e.g., above $16$), enabling a reduction in communication bandwidth and edge device compute complexity. 

\subsection{CIFAR10 - Non-IID}

Compared to the CIFAR10 IID case of Section \ref{cifariid}, we observe a performance drop in terms of overall accuracy as expected; however, even in these challenging non-IID conditions, we observe a convergence of the network test accuracy as the rank $k$ is increased further. This indicates again that using full-rank matrices is not always necessary since the final accuracies converge to similar values as $k\geq8$. Hence, keeping $k$ low is again possible within this context, enabling a reduction in communication bandwidth and edge device compute complexity.
\begin{figure}[htbp]
    \centering
    \includegraphics[width=0.5\textwidth]{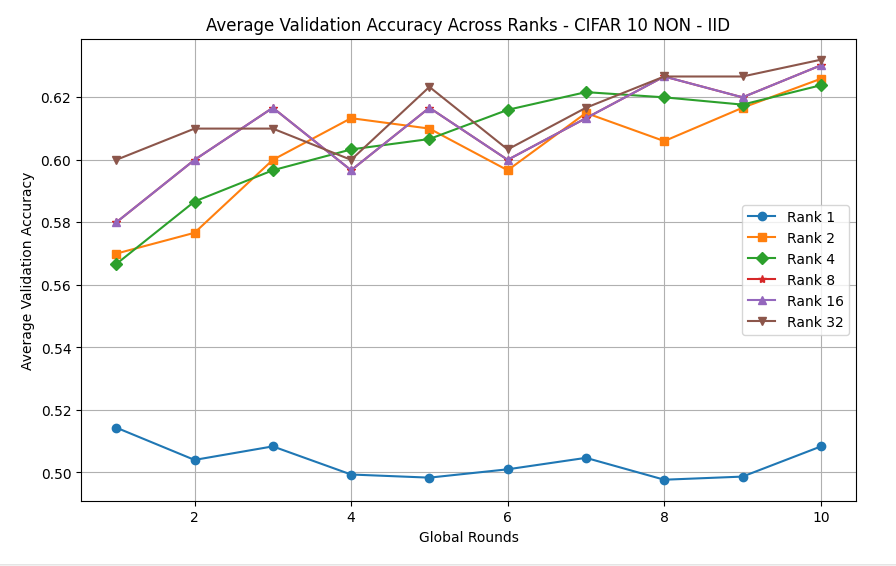} 
    \caption{\textit{\textbf{CIFAR10 NON-IID:} Test accuracy in function of the global FL round.}}
    \label{fig5}
\end{figure}

\subsection{Impact of rank $k$ on the communication bandwidth and edge device compute complexity}
\label{computecomp}
\begin{figure}[htbp]
    \centering
    \includegraphics[width=0.49\textwidth]{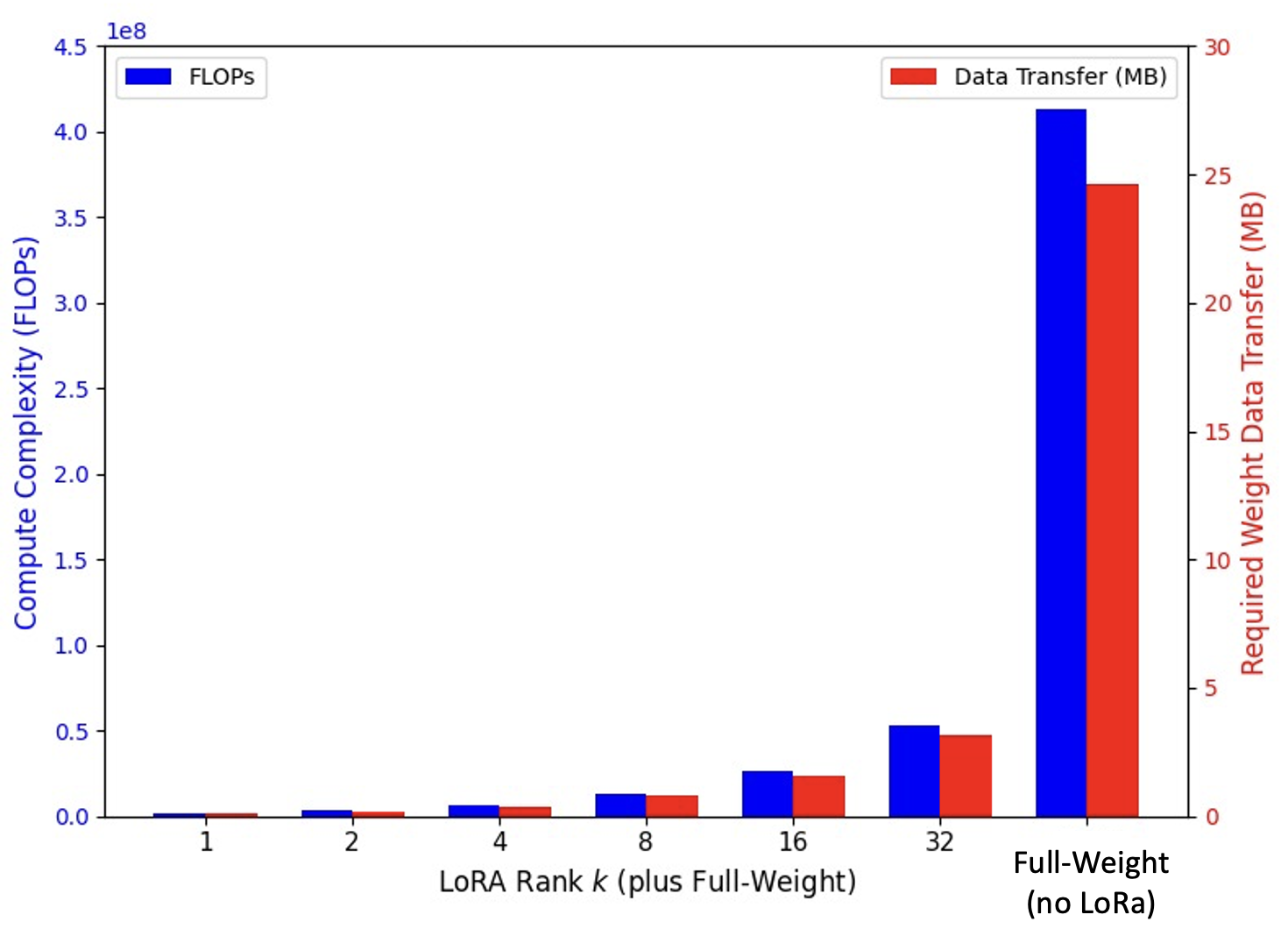} 
    \caption{\textit{\textbf{Edge device compute complexity (FLOPs) and weight communication bandwidth in function of the LoRa rank $k$.} Reducing the rank $k$ leads to an exponential reduction in compute complexity (blue bars) and communication bandwidth (red bars). The compute complexity and communication bandwidth for the case where LoRa is not used (i.e., full-rank weight matrices) are also reported as the last bars in the figure.}}
    \label{fig7}
\end{figure}

To further illustrate the usefulness of our approach, Fig. \ref{fig7} shows the compute complexity and the edge-server weight transfer communication bandwidth required for our one-shot Siamese network in function of the LoRa rank $k$. The compute complexity is evaluated as the number of Floating Point Operations (FLOPs) required during the LoRa weight training in the edge device, while the edge-server communication bandwidth is computed as the amount of weight data in Mega-Bytes (MB) that needs to be transferred back and forth between the edge and the FL server.

Fig. \ref{fig7} shows that a reduction in the rank $k$ leads to an \textit{exponential reduction} in edge device compute complexity and communication bandwidth. This clearly demonstrate the usefulness of using LoRa adapters within FL scenarios, which provide a scalable way to increase or reduce both the required communication bandwidth and the edge device compute complexity depending on the model accuracy requirements. For example, if above a certain rank value $k > k^*$, a model does not present significant gains in terms of accuracy, it is possible to keep the rank of its LoRa modules to $k=k^*$ in order to attain a good compromise between model accuracy, and communication bandwidth and compute complexity. 

For example, it can be seen in Fig. \ref{fig2} that using a rank of only $k=8$ is sufficient for attaining $97\%$ recognition accuracy within the MNIST - IID case. Hence, Fig. \ref{fig7} shows that using $k=8$ instead of $k=32$ leads to significant reductions of more than $\sim 3.6 \times$ in terms of FLOPs and more than $\sim 4.25\times$ in terms of required communication bandwidth. These gains are even more pronounced when comparing our LoRa-FL method to the use of conventional DNN models used in most FL scenarios that do not make use of LoRa modules. In this case, compared to a similar Siamese architecture \textit{without} LoRa modules (i.e., with full-rank weights), Fig. \ref{fig7} shows that our method with $k=8$ enables a drastic reduction of more than $\sim 31 \times$ in terms of FLOPs and more than $\sim 35 \times$ in terms of required communication bandwidth. This discussion clearly illustrate the usefulness of using LoRa modules within FL contexts.  

In sum, our proposed use of LoRa modules within FL scenarios enables a control of the trade off between model accuracy, communication bandwidth and edge device compute complexity, making it a highly attractive method for the design of efficient yet accurate federated learning systems.


\section{Conclusion}
\label{conc}
This work presents a novel federated learning framework that integrates LoRA adapters within a one-shot Siamese network architecture to achieve both communication efficiency and computational savings. By performing training on compact low-rank (LoRa) adapter modules, our method significantly reduces the number of parameters that need to be transmitted and processed during each global round, thereby alleviating the bottlenecks inherent in traditional full-model federated learning. 
Our experimental evaluations on both the MNIST and CIFAR‑10 datasets demonstrate that by reducing the rank of the LoRa modules within the network, the proposed approach can achieve a similar accuracy compared to the use of higher-rank weight matrices while drastically lowering the required edge-server communication bandwidth and the edge device computational complexity. These results indicate that low-rank adaptation not only preserves performance but also provides a scalable and resource-friendly solution for deploying FL on heterogeneous edge devices. Future work will further explore ways to mitigate catastrophic interference in the non-IID data contexts, as well as exploring dynamic rank adaptation and the extension of our LoRa-FL method to more diverse and large-scale datasets.

\end{document}